\documentclass[conference]{IEEEtran}
\IEEEoverridecommandlockouts
\usepackage{cite}
\usepackage{amsmath,amssymb,amsfonts}
\usepackage{algorithmic}
\usepackage{graphicx}
\usepackage{textcomp}
\usepackage{xcolor}

\usepackage{hyperref}
\usepackage{url}

\usepackage[utf8]{inputenc}
\usepackage{alphabeta}

\usepackage{booktabs}
\usepackage{caption}
\usepackage{multirow}

\usepackage{verbatim}

\def\BibTeX{{\rm B\kern-.05em{\sc i\kern-.025em b}\kern-.08em
    T\kern-.1667em\lower.7ex\hbox{E}\kern-.125emX}}

\begin{document}

\title{End-to-End Models for Chemical–Protein Interaction Extraction: Better Tokenization and Span-Based Pipeline Strategies   \\
\thanks{Research reported in this paper was supported by the National Library of Medicine of the National Institutes of Health (NIH) under Award Number R01LM013240. The content is solely the responsibility of the authors and does not necessarily represent the official views of the NIH.}
}

\author{\IEEEauthorblockN{Xuguang Ai}
\IEEEauthorblockA{\textit{Department of Computer Science} \\
\textit{University of Kentucky}\\
Lexington, USA \\
xuguang.ai@uky.edu}
\and
\IEEEauthorblockN{Ramakanth Kavuluru}
\IEEEauthorblockA{\textit{Division of Biomedical Informatics} \\
\textit{Dept.~of Internal Medicine, Univ.~of Kentucky}\\
Lexington, USA \\
ramakanth.kavuluru@uky.edu}
}

\maketitle

\begin{abstract}
End-to-end relation extraction (E2ERE) is an important task in information extraction, more so for biomedicine as  scientific literature continues to grow exponentially.  E2ERE typically involves identifying entities (or named entity recognition (NER)) and associated relations, while most RE tasks simply assume that the entities are provided upfront and end up performing relation classification. E2ERE is   inherently  more difficult  than RE alone given the potential snowball effect of errors from NER leading to more errors in RE.
A complex dataset in biomedical E2ERE is the ChemProt dataset (BioCreative VI, 2017) that identifies relations between chemical compounds and genes/proteins in scientific literature. ChemProt is included in all recent  biomedical natural language processing benchmarks including BLUE, BLURB, and BigBio. However, its treatment in these benchmarks and in other separate efforts is typically not end-to-end, with  few exceptions. In this effort, we employ a span-based pipeline approach to produce a new state-of-the-art E2ERE performance on the ChemProt dataset, resulting in  $> 4\%$ improvement in F1-score over the prior best effort. Our results indicate that a straightforward fine-grained tokenization scheme  helps   span-based approaches excel in E2ERE, especially with regards to handling complex named entities. Our error analysis also identifies a few key failure modes in  E2ERE for ChemProt.
\end{abstract}


\section{Introduction}

\begin{table*}[htbp]
\caption{ChemProt relations grouped based on biological semantic  classes. The five bold CPR groups are used in evaluation. \label{tb-classes}}
\centering
\renewcommand{\arraystretch}{1.3}
\begin{tabular}{lcl}
\toprule
\textbf{Group} & \textbf{Eval} & \textbf{ChemProt relations belonging to the group}  \\
\midrule
CPR: 1 & N & PART\_OF   \\
CPR: 2 & N & REGULATOR $|$ DIRECT\_REGULATOR $|$ INDIRECT\_REGULATOR  \\
\textbf{CPR: 3} & Y & UPREGULATOR $|$ ACTIVATOR $|$ INDIRECT\_UPREGULATOR   \\
\textbf{CPR: 4} & Y & DOWNREGULATOR $|$ INHIBITOR $|$ INDIRECT\_DOWNREGULATOR   \\
\textbf{CPR: 5} & Y & AGONIST $|$ AGONIST-ACTIVATOR $|$ AGONIST-INHIBITOR   \\
\textbf{CPR: 6}& Y & ANTAGONIST   \\
CPR: 7 & N & MODULATOR $|$ MODULATOR-ACTIVATOR $|$ MODULATOR-INHIBITOR   \\
CPR: 8 & N & COFACTOR   \\
\textbf{CPR: 9} & Y & SUBSTRATE $|$ PRODUCT\_OF $|$ SUBSTRATE\_PRODUCT\_OF   \\
CPR: 10 & N & NOT   \\
\bottomrule
\end{tabular}
\end{table*}

Although it is amazing to see rapid progress in biomedical research as we saw during the ongoing COVID-19 pandemic, the associated general explosion of peer-reviewed literature in life sciences can be daunting for researchers to keep up with on a regular basis.   As shown by Lu~\cite{lu2011pubmed}, the exponential growth in scientific literature makes it generally untenable to stay abreast of all the exciting outcomes in a field. To mitigate this situation, natural language processing (NLP) methods that automatically extract relational information reported in literature have been on the rise. Popular relational information of this type includes protein-protein interactions (to understand disease etiology and progression), gene-disease associations (to identify potential drug targets), drug-disease treatment relations (to spot off-label usage or assess potential for repositioning), and drug-gene interactions (to design targeted therapies). Normalizing the extracted relations and storing them in a structured database enables researchers to quickly search for existing research outcomes to arrive at new hypotheses and expedite the knowledge discovery process. 

\subsection{Introduction to the ChemProt task}
\label{sec-task-intro}
Toward reliable benchmarking of NLP methods, over the past decade, there has been a general push to create expert annotated datasets that are used in shared tasks and are subsequently made publicly available for the wider community. The BioCreative series is one such popular venue which has led to many public datasets in BioNLP. The ChemProt extraction shared task that was part of the BioCreative VI series~\cite{krallinger2017overview} is a popular task, included in well-known BioNLP benchmarks such as BLUE~\cite{peng2019transfer}, BLURB~\cite{gu2021domain}, and BigBio~\cite{fries2022bigbio}. The task deals with identifying relations between chemical compounds and proteins (gene compounds\footnote{As genes encode and direct protein production, the gene and the associated protein name are typically the same and refer to either entity depending on the context.}) from scientific literature. For instance, consider the sentence: ``Contribution of the \textbf{Na+-K+-2Cl- cotransporter} NKCC1 to \textbf{\color{blue}Cl-} secretion in rat OMCD.'' Experts annotated this sentence with the \textit{subject} chemical entity \textbf{\color{blue}Cl-} to be a \textit{substrate of} the \textit{object} gene entity \textbf{Na+-K+-2Cl- cotransporter}. 
The resulting relation (called ``interaction'' in the ChemProt task) is often expressed as a triple: (\textbf{\color{blue}Cl-}, \textit{substrate of}, \textbf{Na+-K+-2Cl- cotransporter}), where the relation label (here, \textit{substrate of}) is typically called a \textit{predicate}. In the ChemProt shared task, the spans of both the subject and object entity within the sentence were provided and the participants were asked to predict the type of chemical-protein relation between them from a pre-determined set of such predicates shown in Table~\ref{tb-classes}. (The  \textit{substrate of} predicate is coded as \textbf{CPR: 9} in the dataset.) Long chemical names involving non-alphabetic characters along  with overlapping/nested entities complicate named entity recognition (NER) in ChemProt; relation extraction (RE) is also hampered with long sentences with complex syntactic structures. 

The original ChemProt shared task formulation in 2017 and many efforts that used it in the following years (including latest BioNLP benchmarks BLUE, BLURB, and BigBio) assume  that the chemical and protein names were already spotted in the text. That is, the exact locations of the entities within the sentence are disclosed as part of the input and the task boils down to predicting which chemical-protein pairs participate in an interaction. While this non end-to-end (E2E) setting is important to isolate and evaluate the ability to correctly predict interactions when their spans are available, a more realistic setting for end-user applications is when only raw input is provided. To automatically parse literature to obtain new interactions, there is no scope for nicely spotted entities. Hence there has been a rise in E2ERE methods that incorporate NER as part of the RE process. There are few efforts that handled ChemProt in the E2E setting and that is the focus of our manuscript. 

\subsection{Prior efforts on the ChemProt task}
As indicated earlier, several prior efforts on ChemProt are modeled as relation classification, and hence not E2E. Researchers experimented with convolutional/recurrent neural networks~\cite{liu2018extracting} and tree long short-term memory neural networks (LSTMs)~\cite{lim2018chemical}. We also participated in the shared task and used ensembles of SVMs and neural models to achieve the best performance at the time~\cite{peng2018extracting}, which was subsequently improved upon by  Sun et al~\cite{sun2019deep}. 
Recently, Choi et al.~\cite{choi2020extracting} used calibration methods and self-training using contextualized language models to further improve upon prior results.  

Among the few efforts that attempted E2ERE on ChemProt, Luo et al.~\cite{luo2020neural} used a sequence labeling approach using BiLSTMs with a conditional random field (CRF) layer with additional contextualized features to identify entities and used rules to extract relations. Zuo and Zhang~\cite{zuo2022span} developed a span-based method, SpanMB\_BERT, that considers all span representations up to a max length, subsequently determining span types and potential relations between spans with valid types. They also used another span-based model, DyGIE++~\cite{wadden2019entity}, that jointly models NER and RE using span representations built with a novel idea of span graph propagation, where a graph structure is imposed on spans via different heuristics. The most recent effort by Sun et al.~\cite{sun2022mrc4bioer} also conducts an E2E study on ChemProt. However, after carefully perusing their manuscript, they inadvertently appear to evaluate their model only on test sentences that contain at least one relation. As pointed out by Taille et al.~\cite{taille2020let}, this can inflate the eventual performance as sentences that do not contain any relations could have produced false positives. In fact, 83\% of sentences in the ChemProt test set do not lead to any relations.

\subsection{Our contributions}
\label{sec-contrib}
We believe it is important to continue building E2ERE models for biomedicine, especially on existing publicly available datasets. Using a span-based RE method, we do this for the ChemProt dataset  with the following contributions.
\begin{itemize}
   \item While span-based methods help with overlapping and nested entities that are common in ChemProt, tokenization has a major effect on which entities can be captured, no matter how sophisticated the NER model is. For example, the \textbf{Na+-K+-2Cl- cotransporter} span (from Section~\ref{sec-task-intro}) has ``Na+'' and ``K+'' as gold chemical entities besides the full span encoded as a gene name. The popular ScispaCy biomedical tokenizer~\cite{neumann2019scispacy} outputs only two tokens: ``Na+-K+-2Cl-'' and ``cotransporter''. The NLTK tokenizer~\cite{loper2002nltk} results in seven tokens: ``Na'', ``+-'', ``K'', ``+-'', ``2Cl'', ``-'', and ``cotransporter'', and all spans composed of them will still miss the gold spans  `Na+'' and ``K+''. We employ a simpler, more fine-grained tokenizer to ensure that we don't lose many entities in the pre-processing phase. 
    \item Pipeline models typically have an NER model and a separate RE model, where the NER model output is fed to the RE model. Due to the snowball effect of errors in pipelines where NER errors lead to substantial losses in RE performance, they have fallen out of favor in E2ERE, with researchers looking more toward joint extraction methods~\cite{miwa2016end,tran2019neural,eberts2021end}. However, Zhong and Chen~\cite{zhong2021frustratingly} showed that clever pipeline designs, specifically using typed markers for entities, can help achieve better results than joint models. We use their PURE approach~\cite{zhong2021frustratingly} along with different relation-context representations in combination with our tokenization scheme to achieve new state-of-the-art performance for the ChemProt task. 
    \item We analyze different E2ERE error types, those caused by NER errors and those that have to do with the inherent complexity of the ChemProt interaction types.    
\end{itemize}
Although ChemProt is available from the creators of the dataset\footnote{\url{https://biocreative.bioinformatics.udel.edu/news/corpora/chemprot-corpus-biocreative-vi/}}, to enable fair end-to-end comparisons by other researchers, we make the exact spans (based on our tokenization) publicly available along with the associated pre-processing and modeling code: \url{https://github.com/bionlproc/end-to-end-ChemProt}.

\begin{figure*}[htbp]
  \centering
  \includegraphics[width = 1.0\textwidth]{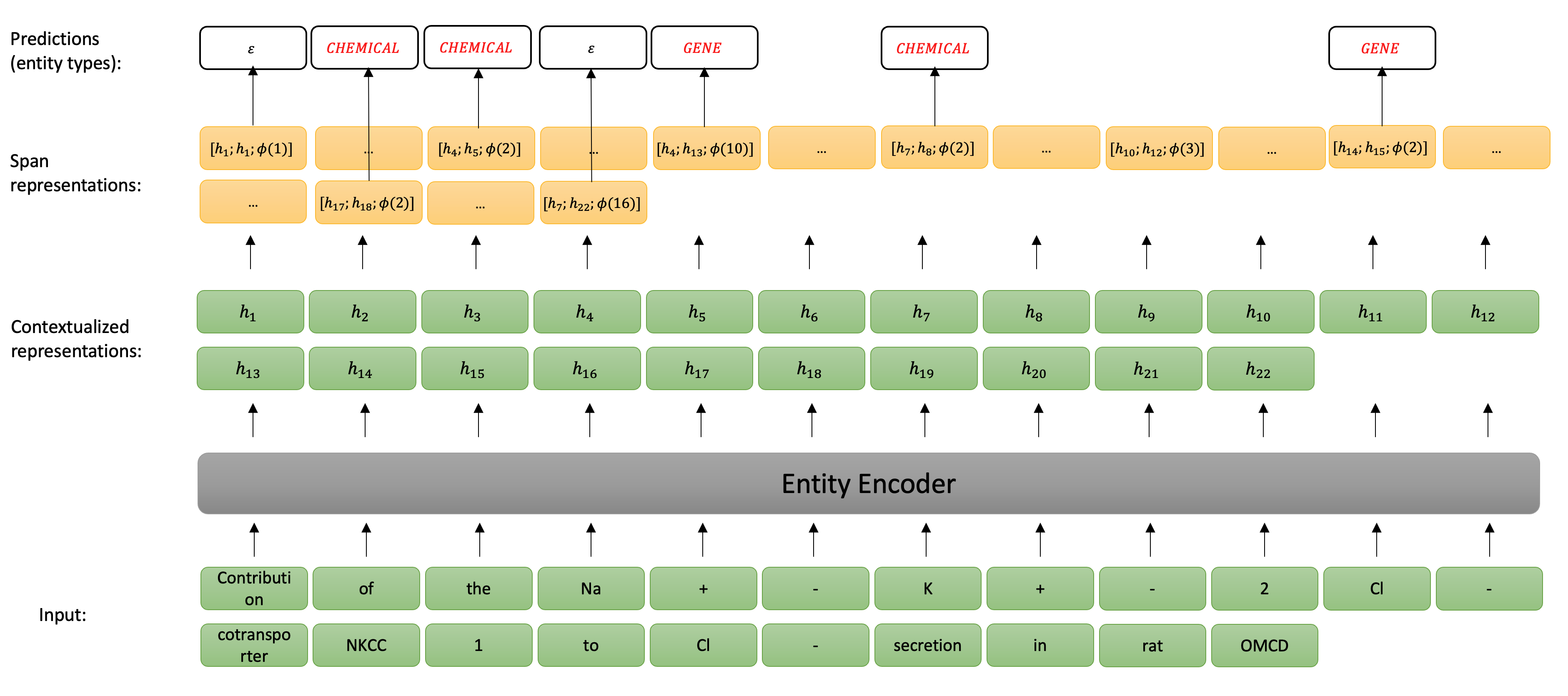}
  \caption{Span-based NER model illustrated with a ChemProt sentence, where for the yellow span representations, the span boundary token embeddings (the $h_i$) are concatenated with the special token embeddings $\phi(l)$ for each $l \geq 1$ for span length. \label{fig-ner}}

\end{figure*}

\section{Methods}

\subsection{The ChemProt dataset and task}

The ChemProt corpus contains 4,966 PubMed abstracts in total: 1,020 training documents, 612 development documents, and 800 test documents. Gold annotations for entities include the exact character start and end positions and entity types: CHEMICAL or GENE. A CHEMICAL and GENE entity can be connected by one of ten relation types indicated in Table~\ref{tb-classes}. However, only five of them are used for evaluation in the task (CPR:3, CPR:4, CPR:5, CPR:6, and CPR:9) because the others are not as interesting~\cite{krallinger2017overview}. The numbers of entities and relations in the training, development, and test sets of ChemProt are shown in Table~\ref{tab-stats}. The original annotation task involves an input sentence that contains the pertinent GENE and CHEMICAL names already annotated. The output is expected to be the relations between all possible CHEMICAL-GENE pairs in it. In the E2E setting, we extend the original task to also spot the entity spans, determine entity types, and extract relations. Overall, the task is at the sentence level. That is, one can assume that the entities will be spelled out directly in the input sentence. However, the full PubMed abstract containing the input sentence is also provided. The annotators are only encouraged to refer to the full abstract when the sentence does not conclusively help with annotating the relations. Hence models are also allowed to look into the full abstract along with the input sentence.

\begin{table}[htbp]
\caption{The counts of entities and relations in the ChemProt dataset. \label{tab-stats}}
\centering
\renewcommand{\arraystretch}{1.3}
\begin{tabular}{lrrr}
\toprule
\textbf{Dataset} & \textbf{Training} & \textbf{Development} & \textbf{Test}  \\
\midrule
CHEMICAL & 13,017  & 8,004 & 10,810  \\
GENE & 12,735  & 7,563  & 10,018 \\ \midrule
CPR: 3 & 768  & 550  & 665 \\
CPR: 4 & 2,254  & 1,094  & 1,661 \\
CPR: 5 & 173  & 116  & 195 \\
CPR: 6 & 235  &  199 & 293 \\
CPR: 9 & 727  &  457 & 644 \\
\bottomrule
\end{tabular}
\end{table}

\subsection{Preprocessing}
\label{sec-preprocess}
Sentence segmentation and tokenization are the first steps in any NLP model and are essential here too. 
We used the Stanza~\cite{qi2020stanza} program for sentence segmentation.
In Section~\ref{sec-contrib}, we demonstrated how ChemProt dataset needs more fine-grained tokenization that does not come naturally with default tokenizers typically available in NLP software. This arises mostly because of the complexity of chemical and gene names that can be a mix of alphanumeric characters mixed with special symbols. 
We use tokenization based on spaces and special symbols using the standard regular expression package in Python~\cite{van2020python}. Precisely, the full regex is  
\begin{verbatim}
"[A-Za-zα-ωΑ-Ω]+|\d+|[^\s]".    
\end{verbatim}
Note that the regex captures groups of  alphabetical or Greek characters as single tokens and treats groups of digits similarly, while considering all other non-space characters as singleton tokens.
For example, with this regex, our running example ``Na+-K+-2Cl- cotransporter'' will be tokenized into ``Na'', ``+'', ``-'', ``K'', ``+'', ``-'', ``2'', ``Cl'', ``-'', and ``cotransporter''. Unlike the ScispaCy  and NLTK tokenizers, these ten tokens fully capture all gold spans, including the full string and the entities ``Na+'' (combination of ``Na'' and ``+'') and ``K+'' (combination of ``K'' and ``+''). Please see how this is different from the extreme tokenization of treating each character as a singleton token, which could lead to too many candidate spans.  

We noticed Wadden et al.~\cite{wadden2019entity} used the ScispaCy method to preprocess the ChemProt dataset but claim to have lost about 10\% of the named entities and 20\% of the relations during the tokenization process.\footnote{Wadden et al.'s pre-processing scripts and results are avaiable online: \url{https://github.com/dwadden/dygiepp}.} Zuo and Zhang~\cite{zuo2022span} use ScispaCy tokens, subsequently split on  `-' and `/' symbols, and disclose that they lost 2\% of entities and relations. Our tokenization  misses only 0.4\% of total entities and 1.37\% of total relations in the training and development datasets, improving over default strategies. 
There are still some entities we cannot  identify even using our fine-grained approach. For example, consider a ChemProt string ``KITD816V'', which will be tokenized to ``KITD'', ``816'', and ``V'' as per our heuristic; however, these tokens cannot be combined in any way to form the gold spans ``KIT'' and ``D816V'' for that string. 

We also found that this preprocessing method is at times helpful in finding obvious annotation errors in the training dataset. For example, in the sentence fragment ``...differentiated with retinoic acid and 12-O-tetradecanoyl-phorbol-13-acetate,'' a clearly incorrect gold entity was provided  as the span ``tinoic acid a''. Since the tokenization we perform will miss this, as we manually examined, we noticed that it was in fact an erroneous annotation while the correct one ought to be ``retinoic acid''. Hence we corrected the gold entity start and end character positions to reflect this new span. This was found in the training abstract with doc\_key = 23194825, in which 47 out of 48 entities were annotated incorrectly by one or two spaces, which were all corrected. Besides this particular example, no other corrections were made in the training or development datasets.

\subsection{The PURE approach: NER and relation models}
\label{sec-pure}
As indicated in the introduction section, we use the Princeton University Relation Extraction (PURE) method~\cite{zhong2021frustratingly} (and some variations). This method is a pipeline of two models, an NER model whose output is passed on to a relation model. PURE's NER model is span-based, in that all possible spans (up to a fixed length) of the input sequence of tokens are considered, one at a time, and tagged with an entity type (including the null type akin to the O tag in IOB tagging scheme). PURE uses a contextualized language model (such as SciBERT \cite{beltagy2019scibert} or PubMedBERT \cite{gu2021domain} in our case) to process the input sequence and obtain contextualized embeddings for each token. For each span, the concatenation of the embeddings of the first and last tokens of the span along with a special token that represents span length is taken as the input to a softmax layer to predict the entity type.
Figure~\ref{fig-ner} illustrates this span-based NER  model with an example ChemProt sentence: ``Contribution of the Na+-K+-2Cl- cotransporter NKCC1 to Cl- secretion in rat OMCD''. For example, $[h_{1}; h_{2}; \phi(2)]$ is the span representation of the candidate entity ``Contribution of''. The figure also shows gold entities ``Na+'' (CHEMICAL), ``Na+-K+-2Cl- cotransporter'' (GENE), ``K+'' (CHEMICAL), ``NKCC1'' (CHEMICAL), and ``Cl-'' (CHEMICAL).

\begin{figure*}[htbp]
  \centering
  \includegraphics[width = 1.0\textwidth]{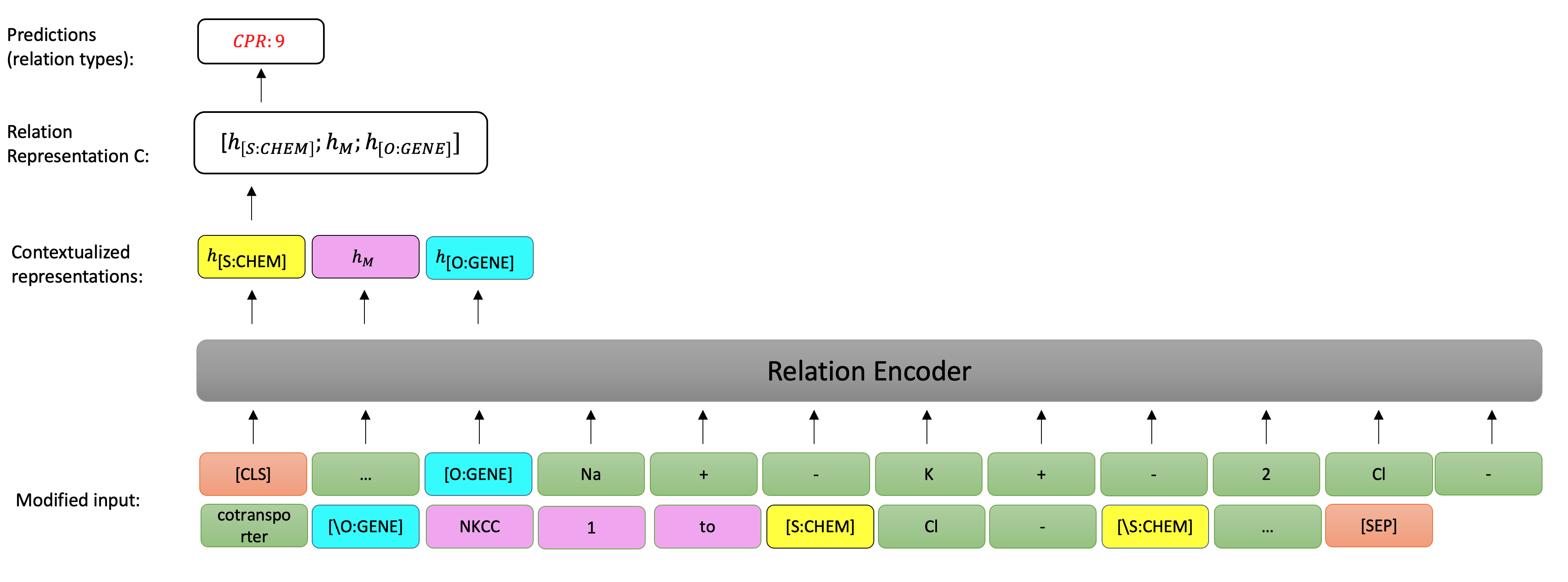}
  \caption{Relation model with an example segment considering the candidate subject ``Cl-'' (CHEMICAL) and object ``Na+-K+-2Cl- cotransporter'' (GENE) enclosed by corresponding entity markers. Here, the relation representation is composed of the start entity marker embeddings along with the intervening text representation ($h_M$). \label{fig-relation-model}}

\end{figure*}

Once the chemicals and proteins are identified, a relation model takes the corresponding spans and predicts if there is a specific relation between each chemical-protein pair. A natural way to do this is to simply take the entity span embeddings from the entity model and use them in combination to predict relations (e.g., via concatenation input to a softmax layer). Note that this would mean that entity span embeddings do not change regardless of which chemical-protein pair is being assessed for a potential relation. Zhong and Chen~\cite{zhong2021frustratingly} argue that entity representations ought to be tailored to each specific pair being considered. To this end, they introduce so called ``entity markers'', which are special start and end tokens that are placed on either side of the entity spans. In the ChemProt task chemicals are always subjects and the genes are objects. Thus we have four special tags \texttt{[S:CHEM]}, 
\texttt{[$\backslash$S:CHEM]}, \texttt{[O:GENE]}, and \texttt{[$\backslash$O:GENE]} that encapsulate the corresponding entity spans in the modified representation. For the candidate chemical-protein pair (``Cl-'', ``Na+-K+-2Cl- cotransporter''), the modified input sequence to the relation model for the earlier example sentence would be:
``Contribution of the \texttt{[O:GENE]} Na+-K+-2Cl- cotransporter \texttt{[$\backslash$O:GENE]} NKCC1 to \texttt{[S:CHEM]} Cl- \texttt{[$\backslash$S:CHEM]} secretion in rat OMCD''\footnote{For convenience we are not showing   the subword WordPiece tokens for the words in this sentence.}. Next, this sequence is input to a pretrained language model to obtain contextualized embeddings for all tokens, including the entity marker tokens. The plan is to use the embeddings of these special tokens as representations of the entities they encapsulate to predict a potential relation between them. In this setting, given how attention modules work in BERT-like language models, it is straightforward to see that the representation of the same chemical (protein) span in a sentence changes based on what protein (chemical) span was being considered to form the candidate pair.  In PURE, these entity marker embeddings are passed on to a two layer feed-forward network with ReLU activations followed by a softmax layers that predicts the relation type. (We use the null $\epsilon$ class to indicate there is no relation along with CPR: 3, 4, 5, 6, and 9 classes from Table~\ref{tb-classes}.) This entity marker scheme to enclose candidate subject and object spans  before passing to a pre-trained language model are demonstrated in Figure~\ref{fig-relation-model}.

The entity model is trained with multi-class cross entropy loss using the estimated probabilities of gold tags corresponding to each span (of length up to $L$). The relation model is trained assuming gold entities are available (given the pipeline approach), with   multi-class cross-entropy using the probabilities estimates of the gold relation type for each candidate chemical-protein pair. At test time, chemicals and proteins from the entity model are passed on to the relation model to infer the interactions.

\subsection{Relation representations}
In Section~\ref{sec-pure}, we conveyed the entity marker embeddings are used to predict the relation type between a chemical and protein. We did not, however, elaborate on how exactly this is to be done. Essentially, a relation representation built on top of the entity marker embeddings is needed. As per the original PURE paper~\cite{zhong2021frustratingly}, the concatenation of entity start contextual embeddings  
\texttt{[S:CHEM]} and \texttt{[O:GENE]}, denoted $r_A = [h_{\texttt{[S:CHEM]}}:h_{\texttt{[O:GENE]}}]$ was passed on to the feed-forward layers for relation prediction. However, we can consider other pieces of evidence too, including $h_{\texttt{[$\backslash$S:CHEM]}}$ and  $h_{\texttt{[$\backslash$O:GENE]}}$, the entity end marker tokens. Furthermore, $h_{[CLS]}$   and the tokens that occur in the middle between the two entities could also contribute signal to the eventual prediction. Inspired by prior efforts (\cite{soares2019matching} and \cite{DBLP:conf/akbc/HoganHKBKVBH21}), we consider different relation representations as shown in Table~\ref{tab-relation-rep}, where $h_M$ refers to an average of all contextualized embeddings of tokens occurring between the chemical and protein spans. $h_M$ is set to the $\mathbf{0}$ vector when there are no tokens between the chemical and protein spans.  

\begin{table}[htbp]
\caption{Relation representations to predict relation types. \label{tab-relation-rep}}
\centering
\renewcommand{\arraystretch}{1.3}
\begin{tabular}{ll}
\toprule
 Notation &  Relation representation based on entity markers\\
\midrule
$r_A$ &  $[h_{\texttt{[S:CHEM]}}:h_{\texttt{[O:GENE]}}]$\\
$r_B$ &  $[h_{[CLS]}: r_A]$\\
$r_C$ &  $[ h_{\texttt{[S:CHEM]}}: h_M: h_{\texttt{[O:GENE]}}]$\\
$r_D$ &  $[h_{[CLS]}: r_C]$\\
$r_E$ &  $[ h_{\texttt{[S:CHEM]}}: h_{\texttt{[$\backslash$S:CHEM]}}: h_M: h_{\texttt{[O:GENE]}}: h_{\texttt{[$\backslash$O:GENE]}}]$ \\
$r_F$ &  $[ h_{[CLS]}: r_E]$ \\
\bottomrule
\end{tabular}
\end{table}

\subsection{Cross-sentence context}

As demonstrated by Zhong and Chen~\cite{zhong2021frustratingly}, cross-sentence context can help language models perform better at E2ERE, especially if pronominal entities are involved.  As indicated earlier, ChemProt task is predominantly designed as a sentence level task, while annotators were allowed to look at the full abstract when needed. So although for many scenarios, cross-sentence signal may not be necessary, it might help in some. To accommodate such situations, we extend each input sentence to the left and right by a fixed number of words denoted by hyperparameters $W_{NER}$ for NER and $W_{RE}$ for RE.

\subsection{Evaluation metrics}
We use precision, recall, and F-scores for both NER performance and E2ERE performance. 
 For NER, predicted entities are considered correct only if entity boundaries and entity types are both correct.
For RE, predicted relations are considered correct only if   boundaries of subject and object entities and relation types are both correct. 
Please note that tokenization already leads to a performance dip (as discussed in Section~\ref{sec-preprocess}) because some entities (and associated relations) cannot be recovered. For ChemProt test set, our tokenization approach loses 0.33\% of entities and 0.98\% of relations. When we determine recall, we consider these missed entities and relations as false negatives, in addition to those arising from recoverable entities/relations.

\section{Experiments and Results}

\subsection{Model configurations and hyperparameters}

In our experiments, we combined the original training   and development datasets in ChemProt to create a combined training dataset and selected 20\% of this  dataset as our new development dataset to tune hyperparameters. This way of splitting the dataset is consistent with other efforts~\cite{luo2020neural,sun2022mrc4bioer}, as the test set is never involved. For pre-trained language models to be used as encoders for the entity and relation models, we used SciBERT, PubMedBERT (Abstracts), and PubMedBERT (Abstracts+PMC). SciBERT was trained on scientific texts: 18\% of papers were from the computer science domain and 82\% were from the broad biomedical domain. PubMedBERT (Abstracts) was trained from scratch (custom vocabulary) using abstracts from PubMed and PubMedBERT (Abstracts+PMC) was trained   using abstracts from PubMed and full-text articles from PubMedCentral.

For all experiments, we trained the entity model for 50 epochs and relation model for 10 epochs.  The batch size is 16 for both NER and RE. We set the context window sizes parameters $W_{NER}=300$ and $W_{RE}=100$  and maximum span length $L=16$ based on empirical assessments with training data. All other model settings (including learning rates) are identical to the PURE model code~\cite{zhong2021frustratingly}. There are very few entities that are composed of more than 16 tokens; since the model fails to capture them, they will be included as false negatives. The performances reported are averages determined across model runs with 5 different seeds.

\subsection{Main results}

Regardless of the base language model used, the NER performances are very similar to each other with F-scores ranging from 90.3 to 91.2 across different seeds\footnote{All performances reported in this paper (including those in the tables) are rounded to the nearest single decimal point.}. Coming to RE performance, the SciBERT relation model F1-scores were in the 65.8 to 66.1 range across different relation representations from Table~\ref{tab-relation-rep}. Relation models with PubMedBERT (Abstracts) and PubMedBERT (Abstracts+PMC) had similar performances across different relation representations with PubMedBERT (Abstracts+PMC) achieving a top RE F-score of 68.8 and PubMedBERT (Abstracts) scoring a top F-score of 69.0. We believe SciBERT did not perform as well considering it was not fully focused on biomedical text. We only show the performances of the relation model across different relation representations for the PubMedBERT (Abstracts) base model in Table~\ref{tab-results}, due to  it's slight edge over PubMedBERT (Abstracts+PMC). From the table, we see that $r_A$, $r_B$, and $r_C$ have the same F-score but slightly different combinations of precision and recall scores; $r_C$ trades-off recall to improve a bit in precision, while $r_A$ achieves the best recall. Including the end tags did not seem to help much ($r_E$ and  $r_F$). 
We compare our best results with prior E2ERE efforts on ChemProt in Table~\ref{tab-compare}, showing a 2\% improvement in NER F-score and over 4\% improvement in RE F-score compared to the prior best results.

\begin{table}[htbp]
\caption{Results with different relation representations  with the PubMedBERT (Abstracts) encoder. \label{tab-results}}
\centering
\renewcommand{\arraystretch}{1.3}
\begin{tabular}{cccc}
\toprule
Relation representations &   P & R & F  \\
\midrule
$r_A$ &69.9 &\textbf{68.3} & \textbf{69.0}  \\
$r_B$ &70.3 &67.8 &  \textbf{69.0}  \\
$r_C$ &\textbf{70.8}& 67.2 &  \textbf{69.0}  \\
$r_D$  & 69.9&67.6 &   68.7 \\
$r_E$  &70.2 &66.9  &  68.5   \\
$r_F$  &70.4 &66.4  &  68.3 \\
\bottomrule
\end{tabular}
\end{table}

\subsection{Ablation of extra context}
We ran a simple experiment where the extra context outside the sentence boundaries designated by   
$W_{NER}=300$ and $W_{RE}=100$ is removed to see if the performance would dip. We perform this ablation for the $r_C$ model with results shown in Table~\ref{tab-ablate}, where the top row includes additional context in both NER and RE models. When the RE context is removed, the performance only dips by 0.4\% in F-score (row 2). When both NER an RE contexts are taken away, the NER performance dips by 0.5\%. Compared to the full context model, the F-score decreases by 0.8\% without the NER and RE contexts. Considering  these dips are all $<$ 1\%, we conclude that the extra context may not have added significant performance boost. This is similar to the small gains observed in the PURE paper~\cite{zhong2021frustratingly}.

\begin{table}[htbp]
\caption{Performances with the $r_C$ model with PubMedBERT (Abstracts) with/without extra context  for NER and RE. \label{tab-ablate}}
\centering
\renewcommand{\arraystretch}{1.4}
\resizebox{\columnwidth}{!}{
\begin{tabular}{lcccccc}
\toprule
\multirow{2}{*}{\textbf{Context window sizes}} & \multicolumn{3}{c}{\textbf{NER}} & \multicolumn{3}{c}{\textbf{RE}}  \\
 & P &  R & F & P & R & F   \\ \midrule
 $W_{NER}$ = 300, $W_{RE}$ = 100 & 91.0 & 90.9 & 91.0& 70.8& 67.2& 69.0\\
 $W_{NER}$ = 300, $W_{RE}$ = 0 &91.0 & 90.9 & 91.0&70.5 &66.8 &68.6\\
$W_{NER}$ = 0, $W_{RE}$ = 0 & 90.5 & 90.4 & 90.4 &69.2 &67.3 &68.2 \\
\bottomrule
\end{tabular}}
\end{table}

\begin{table*}[hbtp]
\caption{A comparison of different end-to-end relation extraction methods for ChemProt. \label{tab-compare}}
\centering
\renewcommand{\arraystretch}{1.4}
\begin{tabular}{lcccccc}
\toprule
\multirow{2}{*}{\textbf{Model}} & \multicolumn{3}{c}{\textbf{NER}} & \multicolumn{3}{c}{\textbf{RE}}  \\
 & P &  R & F & P & R & F   \\ \midrule
Att-BiLSTM-CRF+ELMo \cite{luo2020neural} & 82.5   &   79.8  &  81.1 &   59.5    &  51.2     &  55.1     \\
DyGIE++ \cite{zuo2022span} & 89.7 & 87.6 & 88.7 & 65.4 & 60.5 & 62.9  \\
SpanMB\_BERT \cite{zuo2022span} & 89.3 & 88.3 & 88.8  & 68.0 & 61.5 & 64.6 \\
\textbf{Ours} ($r_C$) & \textbf{91.0} & \textbf{90.9} & \textbf{91.0} & \textbf{70.8} & \textbf{67.2} & \textbf{69.0}  \\
\bottomrule
\end{tabular}
\end{table*}

\section{Error Analysis}

Our main focus in this section is on relation errors in the E2ERE pipeline. Although NER performance is over 90 (in F-score), NER false positives (FPs) and false negatives (FNs) lead to nearly 40\% of RE errors. This is surprising because this implies that the 10\% error rate in NER led to two out of every five RE errors. Potentially, a few missed entities are involved in several gold relations or a few incorrectly spotted entities are leading to many relation FPs. Unfortunately, partial matches lead to both FPs and FNs (for both NER and RE phases). For example, consider the sentence segment --- ``Since this compound retains good \textbf{AChE} inhibitory activity and its hexahydrochromeno[4,3-b]pyrrole moiety is reminiscent of the \textbf{hexahydropyrrolo[2,3-b]indole of physostigmine} (3), \ldots''. Here ``AChE'' is the gold protein and ``hexahydropyrrolo[2,3-b]indole of physostigmine'' is the gold chemical. However, we predict a substring of the chemical, \textbf{hexahydropyrrolo[2,3-b]indole}, leading to an FN for the gold chemical span but also an FP for the substring tagged as the chemical. For this specific example, a gold chemical-protein relation between ``AChE'' and ``hexahydropyrrolo[2,3-b]indole of physostigmine'' is missed due to the chemical FN, causing a relation FN. However, the model actually predicts a new relation between ``AChE'' and the partial matched substring ``hexahydropyrrolo[2,3-b]indole'' leading to a relation FP. Thus, a partial match for a single entity lead to two NER errors and two RE errors. This particular entity is very long (consists of 11 tokens based on our pre-processing approach) and complex and could have been missed by the NER model. There were also occasions where short abbreviated entities are missed by the NER model, especially if they are similar to commonly occurring words or if they are homonymous.

\begin{figure}[htbp]
  \centering
  \includegraphics[width = 0.4\textwidth]{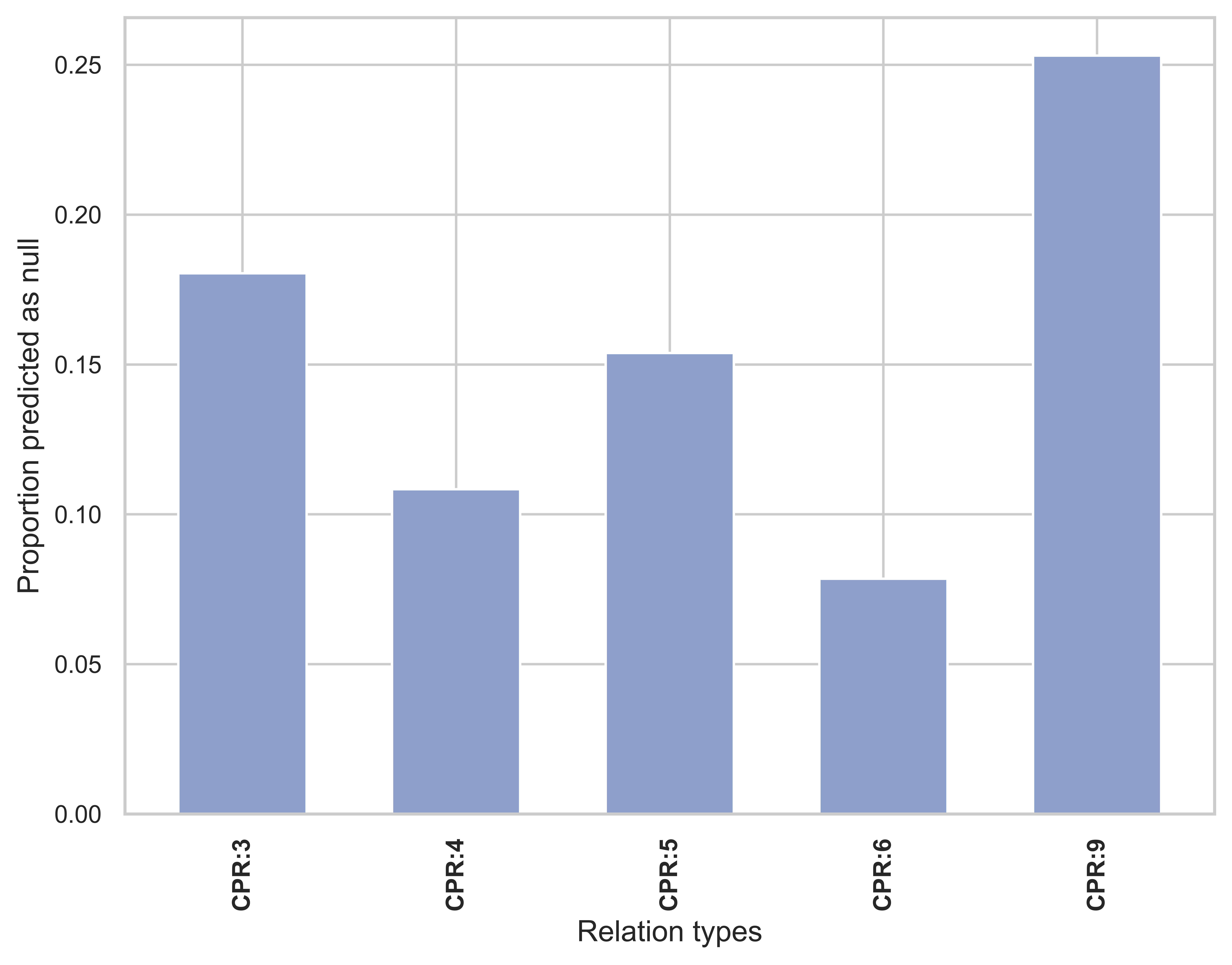}
  \caption{Proportion of gold relations predicted as the null relation\label{fig-cpr9}}

\end{figure}

Next we move on to errors that are not caused by NER errors. These are errors specific to the RE model, where the chemical and protein are correctly tagged by the NER model. 
We begin with FNs where we find that more than 80\% of the errors are when the  model incorrectly predicted a $\epsilon$ (null) relation when the gold label is one of the five valid relations. So this is an not an FN owing to confusion between two different relation types but because of simply not being able to detect any interaction at all. We calculated the proportion of such FN errors for each relation type as shown in Figure~\ref{fig-cpr9}. We clearly see that this happens for one in four CPR:9 (substrate or product of) gold relations. As an example, we had an FN for a gold CPR:9 relation connecting the bold entities in this sentence ---  ``The hypothesis of the present study was that differences among \textbf{dopamine transporter} (DAT) ligands in potency and effectiveness as a positive reinforcers were related to potency and effectiveness as \textbf{DA} uptake inhibitors.'' We speculate here that the substrate relation is not clearly asserted and is rather implied as part of long complicated sentence via an indirect relationship between the chemical and protein. We believe, such long sentences with nuanced expressions may have caused these FNs for CPR:9.

\begin{figure}[htbp]
  \centering
  \includegraphics[width = 0.4\textwidth]{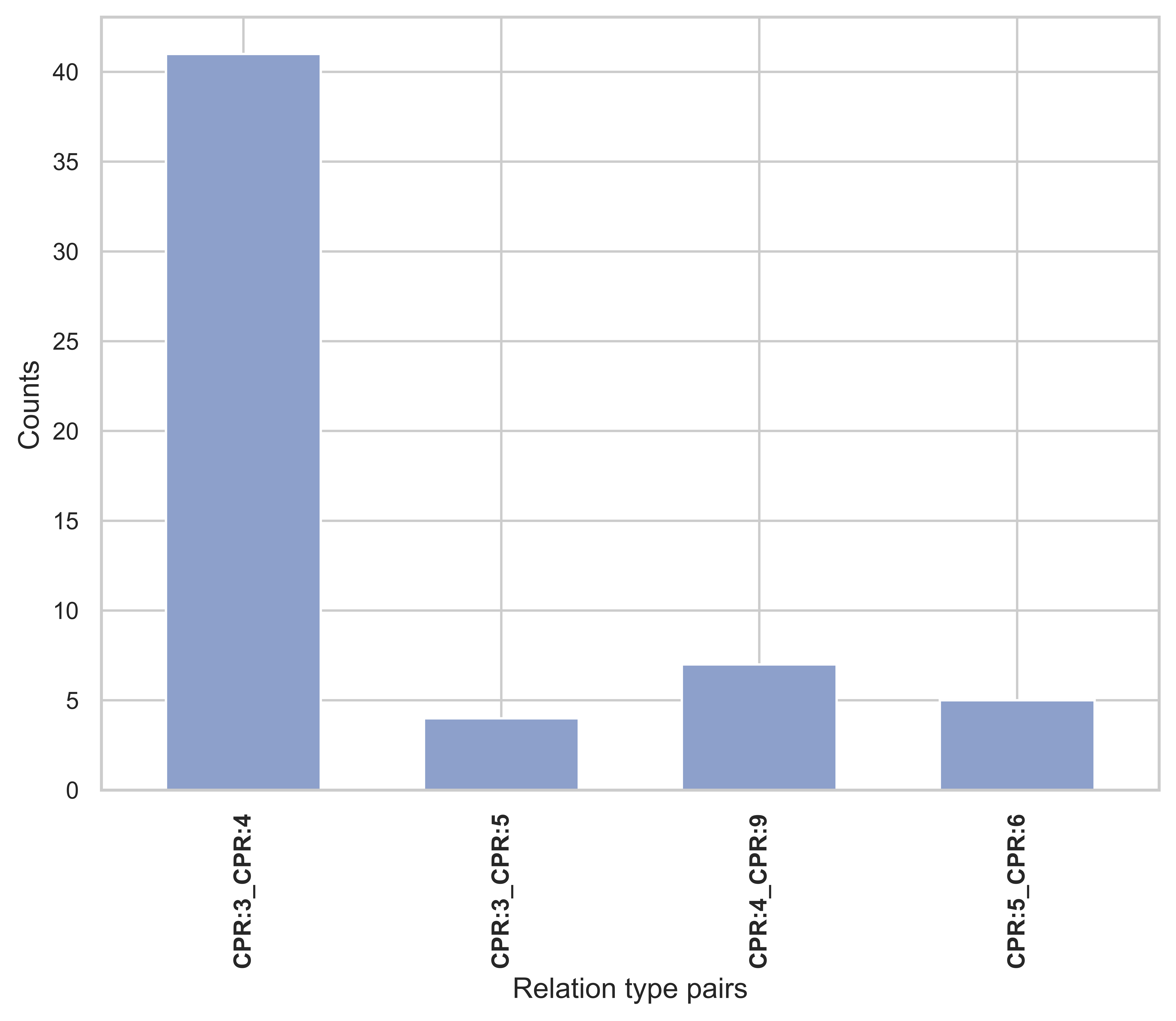}
  \caption{Counts of relation type pairs that are often confused by the model \label{tab-confuse}}

\end{figure}

\begin{figure}[htbp]
  \centering
  \includegraphics[width = 0.4\textwidth]{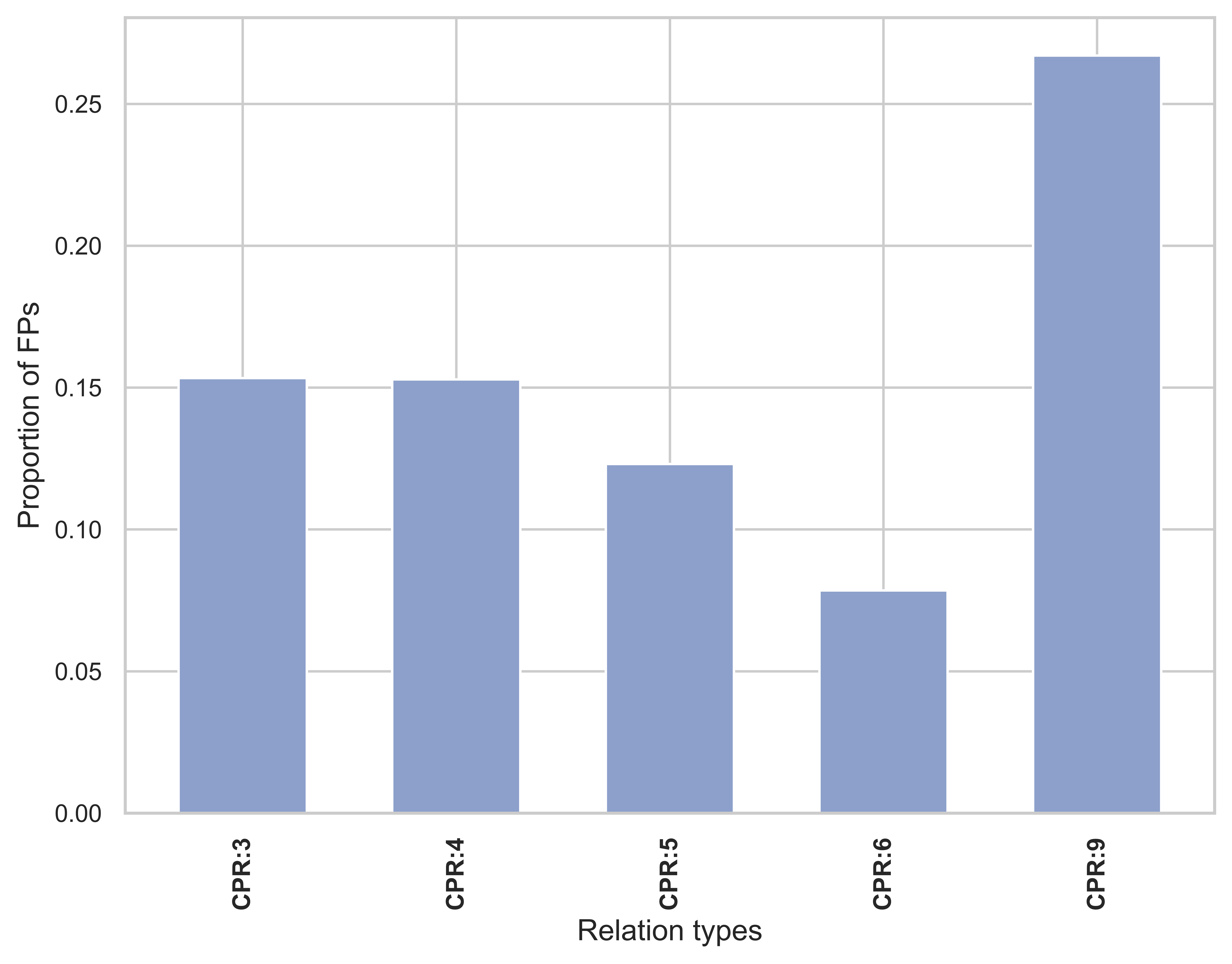}
  \caption{Proportion of FPs in predictions for each relation type \label{tab-fp}}

\end{figure}

A remaining small proportion of FN RE errors is when the model predicted a different non-null relation than the gold relation, the case where both an FN and FP are created. We counted all such errors and found this only happened between certain relation type pairs. We show our results in Figure~\ref{tab-confuse} where the most often confused relations are CPR:3 (upregulator or activator) and CPR:4 (downregulator or inhibitor). This type of errors correspond to what Sun et al.~\cite{sun2019deep} found in their DS-LSTM model too. 
Consider the following sentence where the gold relation was CPR:3 --- 
``Reductions in striatal dopamine and \textbf{tyrosine hydroxylase} content were also less pronounced with \textbf{EHT} treatment.'' The model predicted this as CPR:4. While the phrases ``reductions in'' and ``less pronounced'' in isolation may indicate an inhibitor interaction, the double negative that they induce together seems to indicate an activator link. Since upregulation and downregulation share similar term usage regarding regulation, the model has a hard time telling them apart especially when complex constructs are used as shown in the example.  
We also analyzed the relative proportions of FPs arising from each relation type and found that a quarter of relations predicted as CPR:9 (substrate or product of) are actually FPs (Figure~\ref{tab-fp}). 

\section{Concluding Remarks}
Chemical-protein interactions are central to drug mechanisms leading to both therapeutic potential and side effects. The ChemProt shared task set out to create an NLP benchmark for this important task and since its introduction, most attempts have treated the task as a relation classification problem where the entities are already annotated. Very few attempts were made to address the ChemProt extraction task in an end-to-end manner. In this paper, we improve over prior state-of-the-art in E2ERE for ChemProt using a span-based pipeline approach that additionally uses entity markers in the RE step. We also employ a fine-grained tokenization scheme that retains the ability to extract more entities than the default tokenizers in standard NLP packages. Our improvements are substantial enough (4.4\% in F-score) to have not resulted purely from better tokenization schemes, because the prior best result's tokenization scheme loses 2.0\% of relations due to tokenization while we lose 1.37\%. Ablation experiments show that the extra sentential context adds $<$1\% in performance. 

Although our model improves over prior best scores to a nontrivial extent, the final F-score is still $< 70\%$, which is still 20 points away from 90\%, when models are typically considered powerful and nearing human level performance.   Error analyses showed that long entity spans could hurt NER performance, which stands at 91\%. Since 40\% of RE errors are due to NER errors, the last mile gains in NER performance could greatly improve the E2ERE scores. The substrate relation type (CPR:9) is involved in many FNs and FPs compared to other types and needs to be examined more carefully to  potentially design customized strategies for that type. While   fine-grained tokenization helped the NER step, it could have hurt the RE model because too many tokens within each entity may not capture the semantic representation of the span compared with using fewer but longer tokens. Once the NER step is completed, reverting back to a simpler tokenization scheme could help the RE model better leverage semantic priors in the base language models. Models based on generative approaches such as BioGPT~\cite{luo2022biogpt} (\url{https://huggingface.co/docs/transformers/model_doc/biogpt}) and BioMedLM (\url{https://huggingface.co/stanford-crfm/BioMedLM}) could also be adapted to improve the end-to-end performance and need further exploration.

\vspace{12pt}

\bibliographystyle{IEEEtran}
\bibliography{IEEEabrv, ref}

\end{document}